# Hierarchical Classification System for Breast Cancer Specimen Report (HCSBC) - an end-to-end model for characterizing severity and diagnosis

Thiago Santos, MS[1], Harish Kamath, BS[1], Christopher R. McAdams, MS[5], Mary S. Newell, MD[5], Marina Mosunjac, MD[5], Gabriela Oprea-Ilies, MD[5], Geoffrey Smith, MD[5], Constance Lehman, MD[5], Judy Gichoya, MD[2], Imon Banerjee, Ph.D[2,4], Hari Trivedi, MD,[2,*]

[1]Emory University, Department of Computer Science, Atlanta, Georgia, USA; [2]Emory University, Department of Radiology, Atlanta, Georgia, USA; [3]Mayo Clinic, Phoenix, Arizona, USA; [4]Arizona State University, School of Computing and Augmented Intelligence, Tempe, Arizona, USA; [5]Emory University, Department of Pathology, Atlanta, Georgia, USA; [*] equal contribution.

**Abstract**

*Automated classification of cancer pathology reports can extract information from unstructured reports and categorize each report into structured diagnosis and severity categories. Thus, such system can reduce the burden for populating tumor registries, help registration for clinical trial as well as developing large dataset for deep learning model development using true pathologic ground truth. However, the content of breast pathology reports can be difficult for categorize due to the high linguistic variability in content and wide variety of potential diagnoses ( > 50). Existing NLP models are primarily focused on developing classifier for primary breast cancer types (e.g. IDC, DCIS, ILC) and tumor characteristics, and ignore the rare diagnosis of cancer subtypes. We then developed a hierarchical hybrid transformer-based pipeline (59 labels) - Hierarchical Classification System for Breast Cancer Specimen Report (HCSBC), which utilizes the potential of the transformer context-preserving NLP technique and compared our model to several state of the art ML and DL models. We trained the model on the EUH data and evaluated our model's performance on two external datasets - MGH and Mayo Clinic. We publicly release the code and a live application[1] under Huggingface[2] spaces repository (link).*

## 1 Introduction

Breast cancer is the most common cancer diagnosis in women and the second leading cause of cancer deaths in U.S. women[3]. Each year, over 1M women are biopsied for the detection of only 250,000 breast cancers[4]. During this process, pathologists record highly descriptive observations of cells, organs, and tissue specimens in unstructured and/or semi-structured pathology reports. These reports contain immense quantities of relevant information, which is critical to advance dataset development for cancer research and population registries for cancer epidemiological studies. However, tracking of breast cancer pathological outcomes is currently a largely manual process wherein a human readers codes pathology diagnoses into a database by praising the reports which is a time consuming, expensive, and potentially error-prone task. Our previous work in breast pathology used an online NLP model to extract binary cancer versus benign diagnoses from free-text pathology reports[5], however there is significant value in extracting more granular individual diagnoses from pathology reports. Ultimately, these diagnoses can be used to automatically fill the population research databases or cancer registries.

Automated classification of cancer pathology reports is an active area of research that utilizes machine learning (ML) and deep learning (DL) approaches to extract information from unstructured reports and reduce the burden on human annotators[6–8]. Due to the high linguistic variability in content and wide variety of potential diagnoses, the content of breast pathology reports can be difficult for machines to process. The complexity of pathology ontology hierarchy, clinical diagnoses interspersed with complex explanations, different terminology to label the same cancer, synonymous and ambiguous terms, and multiple diagnoses in a single report are just a few of the challenges. Several studies have explored the use of machine learning and NLP-based methodology for automated cancer information extraction from pathology reports[9]. Earlier studies focused mostly on rule-based techniques and heuristic methods such as dictionary-based and pattern matching methods. Buckley et. al.[10] elaborated a combination of rules for extracting cancer entities (invasive ductal cancer (IDC), invasive lobular cancer (ILC), ductal carcinoma in situ (DCIS), atypical ductal hyplerplasia (ADH), lobular carcinoma in situ (LCIS), and usual ductal hyperplasia (UDH) and its variations, from pathology reports.Nguyen et. al.[11] designed a set of rules in combination with Medtex[12] and SNOMED[13] concepts to identify if malignancy is presented in the pathology report. More recently, machine learning and deep learning approaches have been implemented. Yala et. al[14] presented a boosting classifier using weak learners to identify tumor features. Shang Gao et. al.[15] applied transformer architecture[16] and its attention mechanism for detecting site, laterality, behavior, histology, and tumor grade, respectively. Waheeda Saib et. al.[17] proposed a hierarchical Convolutional Neural Network (CNN) model to automate the classification of breast pathology reports into relevant International Classification of Disease for Oncology (ICD-10) codes. Finally, Mohammed Alawad et. al. developed a single end-to-end multi-task CNN model to extract the primary site, histological grade and laterality from unstructured cancer pathology text reports by using a cross-stitch method to train a word-level CNN, demonstrating how related information may be utilized to create shared representations.[18,19]

Recent advancements in attention based Encoder-Decoder architectures[16,20–23] can be leveraged to improve comprehension of contextual relationships in pathology text mining. The Bidirectional Encoder Representations from Transformers (BERT)[24] is a contextualized language representation model based on a multilayer bi-directional encoder, where the transformer network uses parallel attention layers rather than sequential recurrence to parse the context of the text. Therefore, BERT models can represent words or sequences in a way that captures the contextual information, causing the same sequence of words to have different representations when they appear in different contexts. While there is an increasing interest in developing language models for specific clinical domains, the current tendency tends to favor fine-tuning general transformer models on specialized corpora rather than building models from the ground up with specialized vocabulary[25–27]. However, in domains requiring specialized terminology, such as pathology, these models often fail to capture important medical words in the vocabulary and thus produce sub-optimal performance. One of the primary reasons for this constraint is that many transformer-based models employ WordPiece[28] for unsupervised input tokenization, which is a technique that relies on a predetermined set of Word-Pieces[28]. The word-piece vocabulary is designed to include the most commonly used words or sub-word units, with any new terms (out-of-vocabulary) being represented by frequent sub-words. Although WordPiece was designed to handle suffixes and complex compound words, it frequently fails when it comes to domain-specific lengthier phrases. As example, while ClinicalBERT[26] successfully tokenizes the word "endpoint" as ['end', ##point], it tokenizes the word "carcinoma" as ['car', '##cin', '##oma'] in which the word has lost its actual meaning and replaced by some non-relevant words such as "car".

In this work, we tackle these challenges by developing a hybrid hierarchical NLP pipeline based on a domain-specific PathologyBERT[5] language model for inferring multiple breast pathology diagnoses and disease severity categories from unstructured reports. To our knowledge, we are the first to propose a hierarchical model based on disease severity and clinical management for breast cancer. In addition, this is the first work that applies a domain specific PathologyBERT Language Model[5] to automatically extract disease severity from histopathology reports. We test the effectiveness of the model against several state-of-the-art text classification techniques using a local annotated dataset consisting of 6,681 (six thousand six hundred eighty-one) specimens from unstructured histopathology reports and validate on 926 reports from two external sites. The model classifies each specimen part from a report as containing one or more of 59 individual benign and malignant pathologic diagnoses (invasive ductal carcinoma, lobular carcinoma in situ, radial scar, fibroadenoma, etc.) grouped into 7 severity categories (invasive breast cancer (IBC), in situ breast cancer (ISC), high risk lesion (HRL), borderline lesion (BLL), non-breast cancer (NBC), Benign, and Negative.

The contributions of this study can be summarized as follows. First, we developed a custom ontology to label the most important clinical diagnoses from unstructured breast pathology reports. We then developed a hierarchical hybrid transformer-based pipeline - Hierarchical Classification System for Breast Cancer Specimen Report (HCSBC), which utilizes the potential of the transformer context-preserving NLP technique. Next, we compared our model to several state of the art ML and DL models and evaluated our model's performance on two external datasets. Lastly, we developed a user-friendly graphical user interface (GUI)[1] for our system to upload data and obtain prediction results, and we publicly released the interface and HCSBC model in the Huggingface[2] spaces repository[1]. All model code and weights are released for public use and evaluation[29].

The remainder of this paper is organized as follows. Section 2 reviews related work in the literature on existing NLP methods for automated information extraction (IE) of pathology reports. Section 3 presents the details of the methodology, including the overall framework of the proposed model, detailed analysis of the dataset and the developed annotation tool. Section 4 describes the experiment design and analyzes the results of experiments. Finally, Section 5 concludes this paper and sets forth future work.

## 2 Methods

Figure 1 presents the core processing blocks of our HCSBC pipeline where we formulated the problem as a multi-class classification task. To tackle class imbalance, multi-class prediction, large number of labels, and linguistic variance, we designed a two-level hierarchical model for first predicting cancer severity, and then predicting the individual diagnose(s) within each report. Inspired by Yang et. al.[30], we formulate the task as a hierarchical document classification. As input, the system receives the sequence of tokens describing the report. In the first level of the hierarchy, a transformer-based PathologyBERT model automatically extracts all pathology severities from each report. For example, if the report contains *ductal carcinoma in situ* and *radial scar* then the pathology severities extracted would be *in situ cancer - ISC* and *high risk lesion - HRL*. In the second level of the hierarchy, the report is fed into respective discriminative model branches for each severity category to generate the final diagnosis prediction. This narrows the prediction for individual diagnoses to only those present in the detected severity categories. Each core processing block is described in the following section.

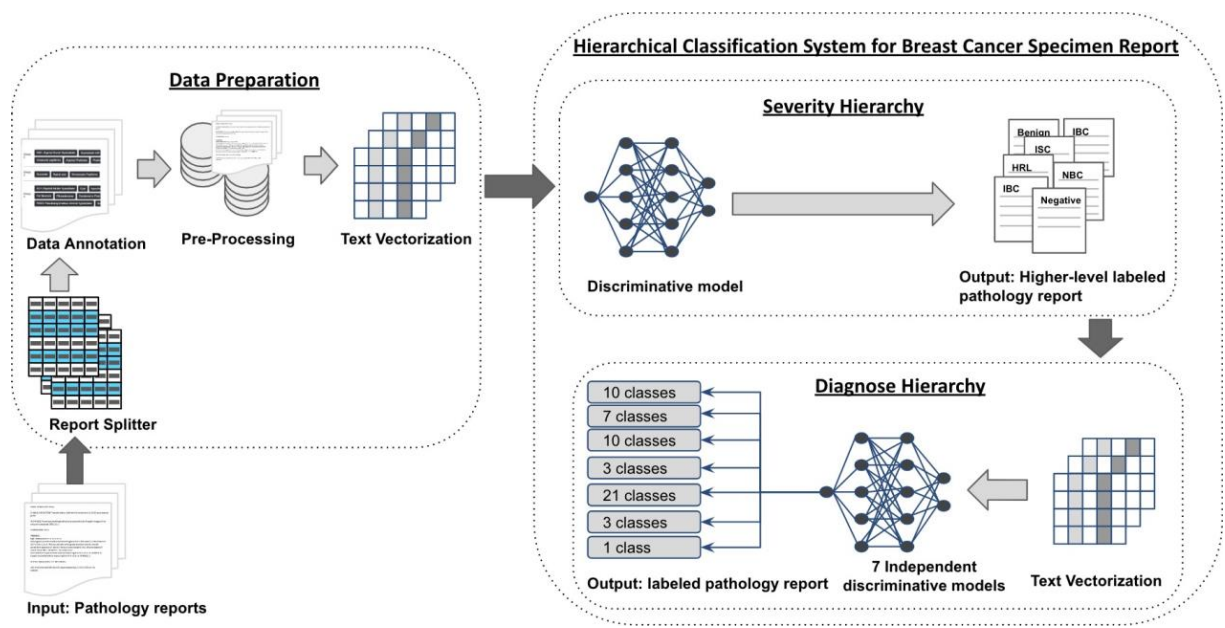

**Figure 1:** An overview of the Proposed Pipeline - core processing blocks are labeled in bold.

### 2.1 Datasets

In order to perform the study and validate the model and results, we collected the following three non-overlapping corpora of breast histopathology reports from three academic healthcare centers.

### 2.2 Internal Dataset

**Corpus 1. Unstructured reports from Emory University Hospital** - With the approval of Emory University Institutional Review Board (IRB), a total of 6681 (six thousand six hundred eighty one) breast cancer pathology reports from unique patients were extracted from four local hospitals at Emory University between 2013-2020. The mean age of the sample is 57 years (+/- 13 years), 38.6% White, 50.5% African American, 4.5% Asian, and 3.9% Hispanic. The final diagnosis section of each reports contains free-text diagnostic information for each specimen, divided by parts denoted by a capital letter and period (A., B., C., etc). The average report length was 42 ± 26 tokens resulting in a corpus size of approximately 2954 tokens.

## 2.3 External Validation Datasets

**Corpus 2. Unstructured reports from Massachusetts General Hospital** - Following IRB approval, 600 breast pathology reports were randomly selected from Massachussets General Hospital (MGH) between 2019-2021 and after dropping irrelavant cases (e.g. non-breast, incomplete), 578 reports were finally evaluated. Regular expressions were used to extract the final diagnosis section from each report and divide this section into individual parts. Each report contained one specimen divided into multiple parts, separated by either a capital letter or number followed by either a period, close parentheses, or colon i.e. 1), A., B:. These were separated by regular expression and incorporated into pre-processing.

**Corpus 3. Unstructured reports from Mayo Clinic** - Following IRB approval, 500 breast pathology reports were randomly selected from Mayo Clinic between 2015 - 2020 across the four sites (La Crosse, WI; Mankato, MN; Rochester, MN; Scottsdale, AZ). 152 number of reports were dropped due to truncated reports, incorrect anatomy, or non-pathology information, yielding 348 reports for model evaluation. Similar to MGH, regular expression was used to split the reports into multiple sections by specimen part.

Figure 5 show the data distribution of severity labels and diagnosis labels for both internal and external dataset. Corpus I is used for training and internal validation of the model. Table 1 present the detailed characteristics of the internal and external datasets.

| Dataset Statistics | Corpus 1 (EUH) | Corpus 2 (MGH) | Corpus 3 (Mayo) |
|---|---|---|---|
| Number of patients | 4,561 | 578 | 348 |
| Number of reports | 6,740 | 578 | 348 |
| **Report level stats** | | | |
| Mean report size | 32 +/- 22 | 41 +/- 31 | 32 +/- 38 |
| Number of Words | 189,337 | 17,850 | 8,806 |
| Number of Tokens | 2,954 | 1,089 | 783 |
| Tokens Exclusive to Corpus | 1,994 | 321 | 126 |
| **Class labels** | | | |
| Number of Invasive breast cancer-IBC | 1,302 | 210 | 67 |
| Number of Non-breast cancer-NBC | 53 | 3 | 0 |
| Number of In situ breast cancer-ISC | 1,192 | 102 | 50 |
| Number of Borderline lesion-BLL | 25 | 0 | 0 |
| Number of High risk lesion-HRL | 1,193 | 129 | 173 |
| Number of Benign-B | 3,684 | 265 | 173 |
| Number of Negative | 1,345 | 42 | 98 |

**Table 1:** Descriptive statistics from corpus 1, 2, and 3.

## 2.4 Breast Pathology Ontology

In consultation with a team of breast pathologists, oncologists, and radiologists, we developed an ontology consisting of 59 diagnoses typically found in breast pathology reports. These diagnoses were divided into seven severity categories - invasive breast cancer (IBC), in situ breast cancer (ISC), high risk lesion (HRL), borderline lesion (BLL), non-breast cancer (NBC), Benign, and Negative largely based on clinical management. A list of diagnoses in each category is provided in the supplement. The ontology was devised to be sufficiently flexible to account for variations in reporting styles, and therefore at least one diagnosis label could be applied to each report. This ontology was used to label breast pathology reports for the local and external validation datasets using the annotation scheme described in Sec. **??**. Experts annotated 6,681 breast pathology reports with the proposed scheme (details in supplement).

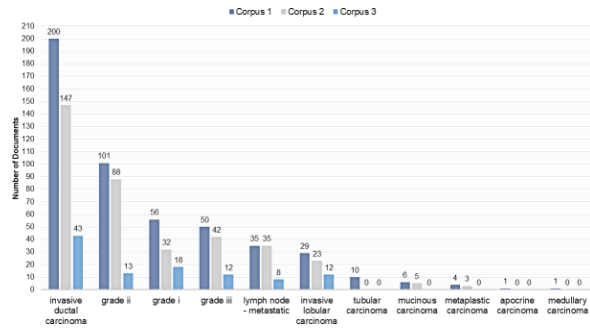
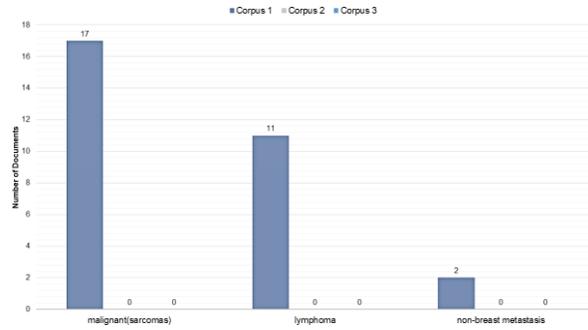
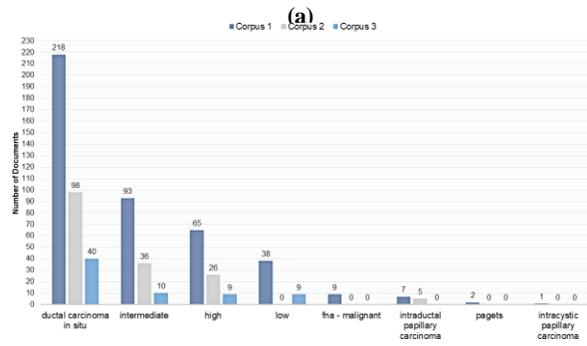
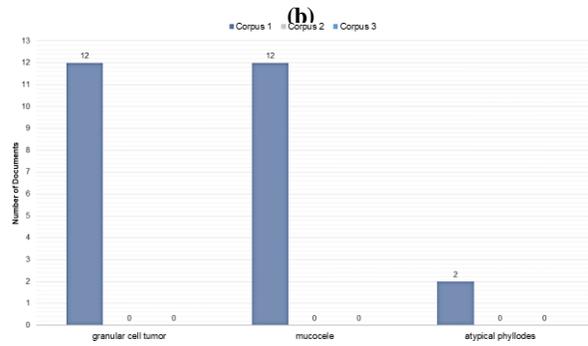
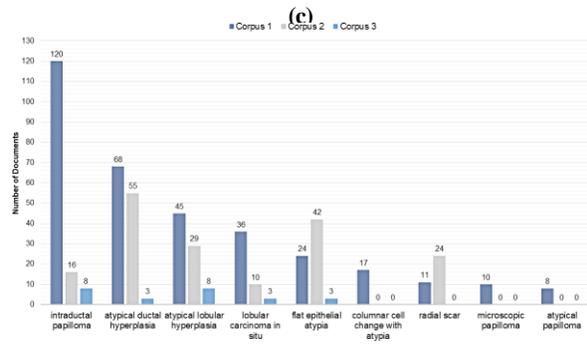
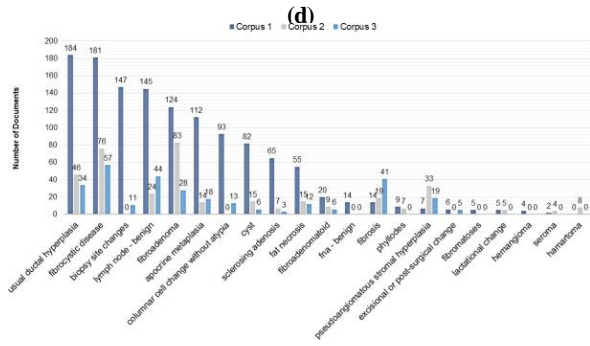

**Figure 2:** Distribution of the class labels presented as bar plot. Class labels are grouped based on the clinical categorization.

### 2.5 Text cleaning & preprocessing

The quality of extracted text features has been demonstrated to be improved by data pre-processing for the NLP models[31]. When building our baseline models, we employ a series of standard pre-processing techniques to minimize the feature space and improve data generalization., including conversion to lowercase and removal of stop words (e.g. 'a', 'an', 'are', etc.) using the Natural Language Tool Kit (NLTK) library[32], and replacing any unique words appearing fewer than five times with an "unknown_word" token. These pre-processing steps were not applied with transformer-based models as they primarily used pre-trained vocabularies without pre-cleaning. For models that did not employ pre-trained techniques, we used a stemming and lemmatization strategy as these models utilize word textual embedding that was trained using the original version of the words.

In addition to standard text pre-processing, we also applied some specific pre-processing techniques for the pathology domain to address numbers and symbols, such as measurements or percentages, which may be important to the diagnosis. All numbers and mathematical symbols were converted into character strings, e.g., '34' was mapped to 'thirty

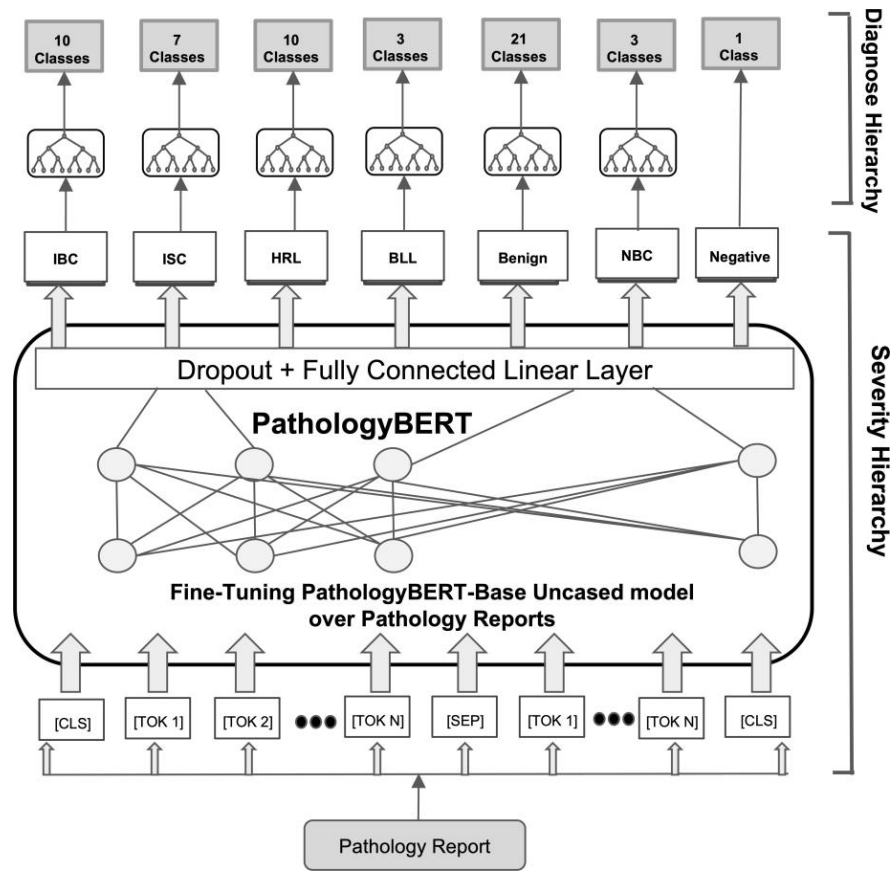

**Figure 3:** Hierarchical Classification System. In the first level of the hierarchy, a PathologyBERT is used to categorize in one (or many) of the seven severity classes. Subsequently, at the second level of the hierarchy, six independent discriminators are trained to extract every cancer diagnose

four' and '%' to 'percentage'. Lastly, the position of the tumor on the breast is indicated by a clock reference with respect to the nipple, however the format was variable throughout the corpus (e.g., 3 o'clock or 03:00). We therefore standardized all clock references to a single format, 1–12, followed by the phrase "o' clock."

### 2.6 Model Design

We propose a two-level hierarchical model as shown in Figure 3: (1) *Severity Hierarchy*: The first is the disease severity classification, which predicts the presence of each disease severity inside of the report (IBC, ISC, HRL, BLL, NBC, Benign, or Negative), and the specific pathology diagnoses contained within an individual severity class; (2) *Diagnosis Hierarchy*: In the second level of the hierarchy, using the gained knowledge of the severity category of the report, we feed each report to its discriminative model(s) branch for specific pathology diagnosis prediction.

#### 2.6.1 Predicting Severity using PathologyBERT: Severity Hierarchy

One of the biggest challenges with existing non-contextual language models[33,34] is handling unknown, out-of-vocabulary (OOV) words, e.g. rare abbreviations, acronyms. While there is a growing interest in developing language models for more specific clinical domains, the current trend appears to prefer fine-tuning general transformer models on specialized corpora rather than developing models from the ground up with specialized vocabulary. However, in fields requiring specialized terminology, such as pathology, these models are not adequate to represent the language space. One of the major reasons for this limitation is because often transformer-based models employ WordPiece tokenizer[35] for unsupervised input tokenization, a technique that relies on a predetermined set of Word-Pieces. The word-piece vocabulary is built such that it contains the most commonly used words or sub-word units and any new words (out-of-vocabulary) can be represented by frequent subwords. Although WordPiece was built to handle suffixes and complex compound words, it often fails with domain-specific longer terms. As example, while ClinicalBERT successfully tokenizes the word "endpoint" as ['end', ##point], it tokenize the word "carcinoma" as ['car', '##cin', '##oma'] in which the word lost its actual meaning and replaced by some non-relevant junk words, such as 'car'. The words which was replaced by the junk pieces, may not play the original role in deriving the contextual representation of the sentence or the paragraph, even when analyzed by the powerful transformer models.

To address this challenge, we extended our previous work to use PathologyBERT- Base Uncased model,[5] which was pre-trained on 340,492 unstructured histopathology reports. As shown with our experimental results, PathologyBERT provides better coverage in breast cancer domain than pretrained BERT models on all diagnoses. Overall prediction accuracy is much higher in PathologBERT (71% accuracy) compared to ClinicalBERT (27%) and BlueBERT (38%). We used the pre-trained PathologyBERT with a fully connected layer on top to fine-tune the whole model to classify disease severity as a downstream task in the pathology domain. Our primary goal was to build a supervised model to first classify our seven severity categories from unstructured breast pathology reports. The implementation is built using the PyTorchTransformers library by huggingface[36].

We fine-tuned the PathologyBERT model using an AdamW optimiser with a learning rate set to $2e^{-5}$ and adopted a weighted binary cross entropy loss to better handle class imbalance. During fine-tuning, we experimented with freezing different layers of the PathologyBERT model and the best results were obtained by only freezing the embedding layer of the architecture. For each experiment, the training was run with an early-stopping patience of 10 epochs. The optimal model was then used to produce the reported results. The maximum sequence length was set at 64 and the batch size at 32. The model architecture and training phase are illustrated in Figure 3. In our preprocessing, a "[CLS]" symbol is added before input texts. The output of highest hidden layer at the "[CLS]" position is taken as a sentence-level feature. Subsequently, a fully connected layer is used to output text classification probability values.

### 2.6.2 Predicting Individual Diagnosis: Diagnosis Hierarchy

In order to formulate second level hierarchy, we constructed a supervised pipeline with 6 independent models (one per severity category) to classify 59 pathologic diagnoses in our ontology. The seventh category (Negative) does not contain any sub-diagnoses so no second-order hierarchy was required. For developing the branched classification models, we experimented with two text different textual embedding: (i) non-contextual - Tf-idf and (ii) contextual - PathologyBERT, and various classification models (Logistic Regression, RandomForest, LSTM. 1DCNN and BERT models). Finally, we identified that RandomForest classifier ourperformed all the other classification pipelines and presented the results using both Tf-idf and pathology BERT embedding.

We also compared the performance of the hierarchical model against different text embedding (tf-idf, word2vec, transformer based embeddings) and machine learning models (Logistic Regression, RandomForest, LSTM. 1DCNN and BERT models) trained on the same dataset but in one single stage for all the 59 classes.

### 2.7 Statistical Validation & Evaluation metrics

We leveraged Corpus 1 to train, test and validate each level of our hierarchy model, by randomly splitting the corpus into 4009 (60%) for training, 1336 (20%) for validation, and 1336 (20%) for testing purpose. We also performed external validation on two datasets from MGH (corpus 2) and Mayo Clinic (corpus 3).

We evaluated accuracy, precision, recall, macro F1 score, and micro F1 score for each model. Because the data is highly class imbalanced, Micro F1 is the standard metric for this task. To further verify the statistical significance of the difference in performance between classifiers, we perform a McNemar's test[37] between the test set predictions of every pair of classifiers. Micro F1 score is calculated as the harmonic average of the precision and recall. Macro F-score, on the other hand, measures classifier performance within each class, and then averages this performance across all classes.

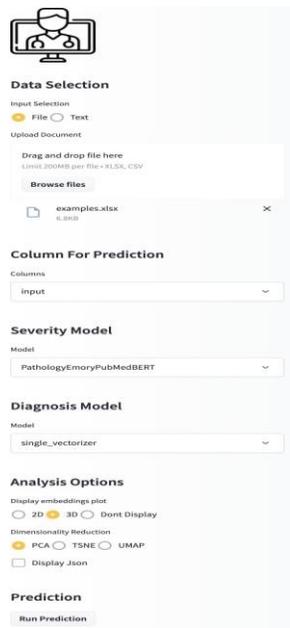

**Figure 4:** A friendly user interface for extracting and predicting: (a) Cancer Severity; (b) Cancer Diagnose. The tool allows the user to run the extraction pipeline from a single specimen input or from a excel spreadsheet with multiple specimen reports.

## 2.8 Interface Design

For extending usability, we developed a user-friendly graphical user interface (GUI) for our system to upload data and obtain prediction results in batch. The interface accepts single specimens input as a string or multiple reports stored in a single file. For each input, the application will return the same outputs as our pipeline - cancer severities and individual diagnoses. Figure 4 illustrates an example of our interface with an auxiliary lymph node specimen. The GUI also allows users to interact with the tool to: i) visualize word importance, ii) visualize BERT sentence embeddings projections, iii) download the predictions in jason format. To allow users to further validate our pipeline and model, we publicly release the code and a live application[1] under Huggingface[2] spaces repository (link).

## 3 Results

### 3.1 Performance on the Severity Hierarchy

Our approach was compared against several state-of-the-art baseline methods for classifying breast cancer histopathology reports in seven severity categories and sixty-one specific diagnoses. The primary aim of the study was to showcase the benefits of integrating contextual embeddings and multi-level hierarchy. Table 2 presents the comparative performance of state-of-the-art text classification models (e.g. traditional XGBoost with tf-idf embedding, transformer models - BioBERT, clinicalBERT) against proposed PathologyBERT model for the classification of the hold-out internal EUH unstructured breast pathology reports into seven severity categories. As observed from the experiments, the PathologyBERT outperforms for every targeted class the traditional and transformer-based model trained on clinical notes and pubmed abstracts, except the Benign and Negative category. This could be due to the overestimation of the negative classes; however the overall performance remain higher for the PathologyBERT. The performance proves the point that optimal coverage of pathology vocabulary achieved better classification performance for the complex task.

| Severity category | XGBoost | | | BioBERT | | | ClinicalBERT | | | PathologyEmoryBERT | | | Support |
|---|---|---|---|---|---|---|---|---|---|---|---|---|---|
| | Precision | Recall | F1 Score | Precision | Recall | F1 Score | Precision | Recall | F1 Score | Precision | Recall | F1 Score | |
| Invasive breast cancer-IBC | 0.94 +/- 0.01 | 0.94 +/- 0.01 | 0.94 +/- 0.003 | 0.96 +/- 0.01 | 0.96 +/- 0.01 | 0.96 +/- 0.008 | 0.97 +/- 0.007 | 0.97 +/- 0.009 | 0.97 +/- 0.006 | **0.97 +/- 0.009** | **0.98 +/- 0.008** | **0.97 +/- 0.006** | 264 |
| Non-breast cancer-NBC | 0.65 +/- 0.1 | 0.33 +/- 0.2 | 0.47 +/- 0.2 | 0.92 +/- 0.05 | 0.83 +/- 0.06 | 0.87 +/- 0.04 | 0.89 +/- 0.05 | 0.89 +/- 0.05 | 0.89 +/- 0.04 | **0.93 +/-0.05** | **0.92 +/- 0.04** | **0.92 +/- 0.02** | 30 |
| In situ breast cancer-ISC | 0.93 +/- 0.01 | 0.90 +/- 0.01 | 0.91 +/- 0.006 | 0.96 +/- 0.01 | 0.94 +/- 0.01 | 0.95 +/- 0.009 | 0.97 +/- 0.009 | 0.95 +/- 0.012 | 0.96 +/- 0.008 | **0.96 +/- 0.01** | **0.96 +/- 0.01** | **0.96 +/- 0.008** | 239 |
| Borderline lesion-BLL | 0.72 +/- 0.04 | 0.65 +/- 0.06 | 0.68 +/- 0.04 | 0.88 +/- 0.03 | 0.73 +/- 0.08 | 0.81 +/- 0.06 | 0.93 +/- 0.004 | 0.85 +/- 0.006 | 0.89 +/- 0.003 | **0.95 +/- 0.042** | **0.87 +/- 0.063** | **0.90 +/- 0.041** | 26 |
| High risk lesion-HRL | 0.90 +/- 0.01 | 0.88 +/- 0.01 | 0.89 +/- 0.004 | 0.94 +/- 0.01 | 0.92 +/- 0.01 | 0.93 +/- 0.01 | 0.93 +/- 0.01 | 0.95 +/- 0.01 | 0.93 +/- 0.01 | **0.94 +/- 0.01** | **0.96 +/- 0.01** | **0.95 +/- 0.009** | 237 |
| Benign-B | 0.85 +/- 0.03 | 0.82 +/- 0.03 | 0.83 +/- 0.002 | 0.90 +/- 0.01 | 0.91 +/- 0.01 | 0.90 +/- 0.008 | 0.91 +/- 0.01 | 0.94 +/- 0.008 | 0.93 +/- 0.006 | **0.91 +/- 0.01** | **0.94 +/- 0.008** | **0.92 +/- 0.007** | 730 |
| Negative | 0.85 +/- 0.05 | 0.81 +/- 0.07 | 0.83 +/- 0.04 | 0.91 +/- 0.01 | 0.85 +/- 0.02 | 0.88 +/- 0.01 | **0.93 +/- 0.015** | **0.985 +/- 0.02** | **0.89 +/- 0.013** | 0.92 +/- 0.01 | 0.85 +/- 0.02 | 0.88 +/- 0.01 | 273 |
| | | | | | | | | | | | | | |
| Accuracy | | 0.78 +/- 0.004 | | | 0.85 +/- 0.008 | | | 0.88 +/- 0.007 | | | 0.89 +/- 0.009 | | 1799 |
| Micro Average | 0.88 +/- 0.005 | 0.82 +/- 0.002 | 0.84 +/- 0.004 | 0.92 +/- 0.007 | 0.91 +/- 0.003 | 0.92 +/- 0.005 | 0.93 +/- 0.005 | 0.92 +/- 0.005 | 0.92 +/- 0.004 | **0.93 +/- 0.005** | **0.93 +/- 0.008** | **0.93 +/- 0.004** | |
| Macro Average | 0.87 +/- 0.03 | 0.71 +/- 0.05 | 0.76 +/- 0.04 | 0.91 +/- 0.05 | 0.88 +/- 0.04 | 0.90 +/- 0.01 | 0.93 +/- 0.006 | 0.91 +/- 0.05 | 0.92 +/- 0.004 | **0.94 +/- 0.006** | **0.94 +/- 0.008** | **0.94 +/- 0.008** | |

**Table 2:** Cancer Severity inference table results from our proposed Hierarchical Classification System on Corpus 1 - EUH Test set. We compared and evaluated several state-of-the-art Transformer models.

In order to show the generalizibility, we also evaluate the PathologyBERT model on the external MGH and Mayo clinic pathology reports (see Table 3). PathologyBERT achieved comparable performance on both internal EUH (89% accuracy) and External: MGH (84% accuracy) and Mayo clinic (86% accuracy). While EUH contains 1.44% borderline lesions, the external datasets do not have any lesion categorized in that category and Mayo clinic corpus does not have any NBC lesion. Other than a significant drop in performance for the Negative cases (0.60 f1-score) for the MGH cohort, the performance stays equally high on the internal and external datasets.

| Cancer Diagnose Severity | Corpus 1 - EUH | | | | Corpus 2 - MGH | | | | Corpus 3 - Mayo | | | |
|---|---|---|---|---|---|---|---|---|---|---|---|---|
| | Precision | Recall | F1 Score | Support | Precision | Recall | F1 Score | Support | Precision | Recall | F1 Score | Support |
| Invasive breast cancer-IBC | 0.97 | 0.98 | 0.97 | 264 | 0.88 | 0.95 | 0.91 | 210 | 0.94 | 0.97 | 0.95 | 60 |
| Non-breast cancer-NBC | 0.93 | 0.92 | 0.92 | 30 | 0.67 | 1 | 0.80 | 2 | - | - | - | 0 |
| In situ breast cancer-ISC | 0.96 | 0.96 | 0.96 | 239 | 0.90 | 0.94 | 0.92 | 102 | 0.96 | 0.99 | 0.98 | 43 |
| Borderline lesion-BLL | 0.95 | 0.87 | 0.90 | 26 | - | - | - | 0 | - | - | - | 0 |
| High risk lesion-HRL | 0.94 | 0.96 | 0.95 | 237 | 0.88 | 0.98 | 0.93 | 129 | 0.72 | 0.95 | 0.82 | 22 |
| Benign-B | 0.91 | 0.94 | 0.92 | 730 | 0.87 | 0.93 | 0.91 | 264 | 0.93 | 0.88 | 0.91 | 181 |
| Negative | 0.92 | 0.85 | 0.88 | 273 | 0.71 | 0.52 | 0.60 | 42 | 0.81 | 0.91 | 0.86 | 96 |
| | | | | | | | | | | | | |
| Accuracy | | 0.89 | | 1799 | | 0.84 | | 749 | | 0.86 | | 402 |
| Micro Average | 0.93 | 0.93 | 0.93 | | 0.87 | 0.93 | 0.90 | | 0.89 | 0.92 | 0.90 | |
| Macro Average | 0.94 | 0.94 | 0.94 | | 0.79 | 0.90 | 0.84 | | 0.87 | 0.94 | 0.90 | |

**Table 3:** Cancer Severity inference table results from Hierarchical Classification System performance on the internal and external datasets: i) Corpus 1 - EUH, ii) Corpus 2 - MGH external validation set, and iii) Corpus 3 - Mayo external validation set.

### 3.2 Performance on the Diagnosis Hierarchy

Our experiments demonstrate that a hierarchical structure that breaks down a complex classification problem into smaller problems helps the model to better locate text segments critical for classification, particularly for reports with rare labels. We tested two versions of the hierarchical model (HCSBC): (1) feed the PathologyBERT embedding to each branch classifier and (2) Use tf-idf to vectorize the input text before feeding to each branch classifier. We also compared HCSBC against the traditional (logistic regression, random forest, XGBoost), neural network (LSTM, 1DCNN) and transformer based (ClinicalBERT, BiomedBERT, CharacterBERT) models. Table 4 summarizes the overall comparative performance of our approach and Fig. 5 represents the label-wise performance as bar plot. It is important to emphasize that when compared to baselines, both versions of the proposed hierarchical model improved the overall performance. HCSBC was able to outperformed all the traditional and nerural network models with significant margin. While HCSBC was able to outperform the best baseline (XGboost with tf-idf embedding) by up to +4 micro F1 score by employing PathologyBERT embeddings on the second stage of the hierarchy, interestingly HCSBC was able to achieve up to +6 micro F1 score when employing a basic tf-idf in the second stage.

| Classifier | Feature Representation | Precision | Recall | F1 Score | P-Value |
|---|---|---|---|---|---|
| Logistic Regression | tf-idf | 0.68 | 0.63 | 0.64 | < 0.0001 |
|  | word2vec | 0.67 | 0.68 | 0.68 |  |
| Random Forest | tf-idf | 0.85 | 0.77 | 0.80 | < 0.0001 |
|  | word2vec | 0.78 | 0.76 | 0.76 |  |
| LSTM | LSTM embedding Layer | 0.75 | 0.71 | 0.71 | < 0.0001 |
|  | word2vec embedding Layer | 0.71 | 0.70 | 0.70 |  |
| 1DCNN | DCNN embedding Layer | 0.73 | 0.71 | 0.71 | < 0.0001 |
|  | word2vec embedding Layer | 0.72 | 0.70 | 0.70 |  |
| XGBoost | tf-idf | 0.89 | 0.79 | 0.84 | < 0.001 |
|  | word2vec | 0.78 | 0.77 | 0.77 |  |
| ClinicalBERT | ClinicalBERT | 0.82 | 0.75 | 0.79 | < 0.0001 |
| BiomedBERT | BiomedBERT | 0.82 | 0.76 | 0.79 | < 0.0001 |
| CharacterBERT | CharacterBERT | 0.84 | 0.77 | 0.80 | < 0.0001 |
| **HCSBC** | PathologyBERT Embeddings | 0.91 | 0.83 | 0.86 | < 0.001 |
|  | **tf-idf** | **0.92** | **0.89** | **0.90** | **Ref. Group** |

**Table 4:** Pathology diagnosis experimental results over randomly hold-out test set of corpus 1. We evaluated our proposed HCSBC pipeline with two distinct feature representations: i) PathologyBERT sentence embeddings representation; ii) tf-idf. Both versions of the HCSBC model uses PathologyBERT embeddings and classifier in the first layer.

From Fig. 5, the model's performance is particularly evident for macro F1-scores, indicating that the model is considerably better at properly classifying cases with very few examples which is crucial for classifying rare histologies. Such performance is preserved when applied to the external datasets - Corpus 2 and 3. Particularly we observer a significant drop in performance for the intermediate category between the internal and external datasets which could be due the uncertainty during the labeling. Another large drop in performance is observed for 'sclerosing adenosis'; however there are only 7 and 3 samples presents in the external cohort.

### 3.3 Error Analysis

To understand the pattern of error during the prediction, we also performed a through error analysis. Table 5 presents example sentences predictions generated by the model in the test set. Errors can be grouped into three main categories - *Wrong Diagnosis*, *Missed Diagnosis*, *and Incorrect Ground Truth*. Errors of *Wrong Diagnosis* were frequently between adjacent classes, for example between *Negative* and *Benign* or *IBC* and *ISC*. In Table 5, row 1 and 2 demonstrate misclassification between *Negative* and *Benign* cases. Missed diagnoses occured most frequently in long reports with complex structure and many diagnoses, however even in these cases the most important diagnoses of IBC, ISC, and HRL were typically retained. Row 3 represents an FNA of a lymph node which was classified as as *FNA - malignant* but represents a metastatic lymph node, indicating invasive cancer. Row 7 represents a long report and difficult case which was divided into subsections in preprocessing. In this case *mucocele* in class *BLL* and *usual ductal hyperplasia* in class *Benign* were missed by the model. *Sclerosis adenosis* was predicted which is not in the original labels. Lastly, there were examples of *Incorrect Ground Truth* in which the original label was incorrect from the annotator. For example a malignant FNA of a lymph node would qualify as *lymph node - metastatic* in class *IBC* rather than *FNA - malignant* in class ISC. Row 4 and 5 represents cases in which the original label was incorrect and the model prediction was correct. Row 6 represents a rare diagnosis of *malignant (sarcomas)* which was not recognized by the model due to rarity of samples.

### 4 Discussion

In this work, we experimented with a wide variety of natural language processing and text mining techniques to process the text of cancer pathology reports to successfully extract relevant information from both internal and external datasets and categorize the reports into structured categories. We propose an effective hierarchical classification model (HCSBC) for the extraction of 59 unique labels corresponding cancer severity and its equivalent diagnose(s). The model tackles the complex pathology language space, clinical diagnoses interspersed with complex explanations, different terminology to label the same cancer, synonymous and ambiguous terms, and multiple diagnoses in a single report. Our proposed hierarchical system is designed as a two-level hierarchical model for first predicting cancer severity, and then predicting the individual diagnose(s) within each report and showed that such two-step processing helps to better focus on the relevant text segment and outperformed all the SOTA models. We publicly release the code and a live application[1] under Huggingface[2] spaces repository (link).

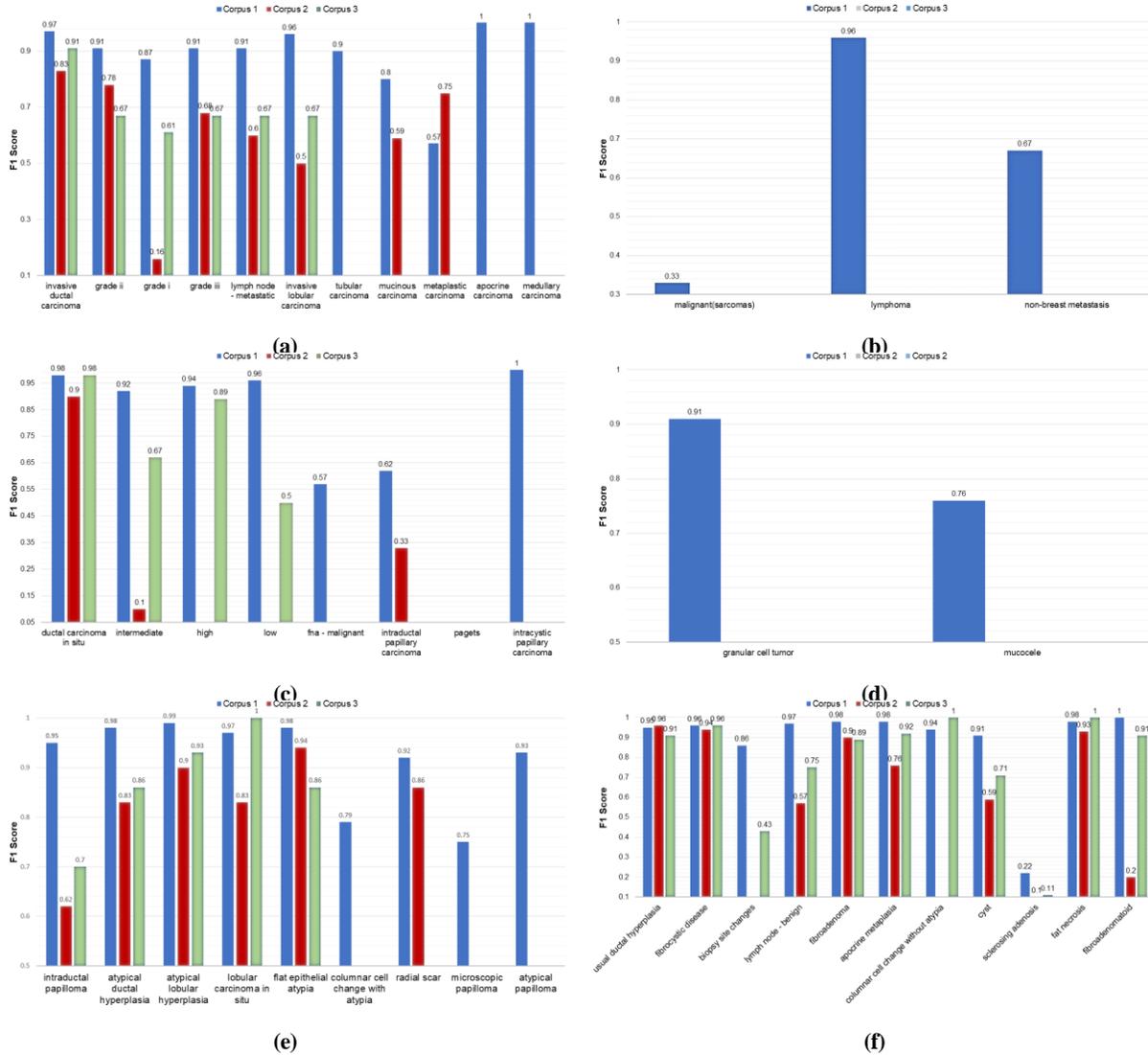

**Figure 5:** Diagnosis hierarchy inference using the HSBSC model on Corpus 1 - EUH test set, Corpus 2 - MGH external validation set, and Corpus 3 - Mayo external validation set. Class-wise performance is measured in-terms of Micro-F1 score. Missing bar represents no presence of that label in the cohort.

The proposed system takes as input text paragraphs describing the report or a csv/xlsx file with a set of reports for batch processing. In the first level of the hierarchy, a transformer-based PathologyBERT model trained on specialized pathology corpora automatically extracts every pathology severity from each report. Subsequently, in the second level of the hierarchy, we feed each report the respective discriminative model branches pertaining to each severity category to generate the final diagnosis prediction (see Fig. 1). This narrows the prediction for individual diagnoses to only those present in the detected severity categories. We assessed our model's performance using a hold-out dataset and two external validation datasets. We demonstrated that our proposed model performed exceptionally well on the internal hold-out test set and it provided an effective performance for the external validations, indicating our model's ability for generalizability.

| Specimen Report | Severity Label(s) | Severity Prediction(s) | Diagnosis Label(s) | Diagnosis Prediction(s) |
|---|---|---|---|---|
| CYST Cyst, Breast, Left : negative for malignant cells. benign ductal cells and macrophages present. | Negative | Benign | Negative | No Prediction |
| BRLUMP Re-excision, right breast. right breast, re-excision: - benign breast tissue with previous surgical site changes, no atypia or malignancy is present. | Benign-B | Negative | biopsy site changes | Negative |
| FNA Fine Needle Aspiration, Axillary Mass, Right : malignant cells present: adenocarcinoma (). | IBC | ISC | lymph node - metastatic | fna - malignant |
| FNA Lt Supraclavicular, Fine Needle Aspiration a: lymph node/soft tissue, left supraclavicular, fine needle aspiration: malignant cells present.metastatic breast adenocarcinom. | ISC | IBC | fna - malignant | lymph node - metastatic |
| BRSTMARUN Right breast wide local excision anterior margin breast, right anterior margin, wide local excision: benign breast tissue negative for melanoma | NBC | Negative | malignant (sarcomas) | Negative |
| SLOS 16 Slides; OSC# S18-24641, DOS-11/28/18 left breast, core biopsy: sarcomatoid malignant neoplasm, please see above. | NBC | IBC | malignant (sarcomas) | No Prediction |
| BRSTWLN Right breast partial mastectomy; short stitch-superior, long stitch-lateral right breast, segmental mastectomy: invasive mucinous carcinoma and microinvasive mucinous carcinoma associated with extensive ductal carcinoma in situ that has a mucocele like lesion pattern ().o the invasive carcinoma measures 4 mm (b7), nottingham grade 1 .o the microinvasive carcinoma measures 1 mm (b11). extensive ductal carcinoma in situ (dcis), low nuclear grade, cribriform type with mucinous and signet ring features and extension into lobules. margins for this specimen (b) are reported as follows (additional margin specimens have been submitted as c-h):o invasive carcinoma is present 6 mm from the superior margin.o dcis is present less than 1 mm from the medial, anterior, superior and posterior margins. no lymphovascular invasion is seen. intraductal papilloma, 11 mm, with usual type ductal hyperplasia and sclerosis. usual type ductal hyperplasia, apocrine adenosis, columnar cell hyperplasia, columnar cell changed calcifications are seen in association with dcis and benign ducts. biopsy site changes. | IBC, ISC, BLL, HRL, Benign-B | IBC, ISC, HRL, Benign-B | columnar cell change without atypia, usual ductal hyperplasia, mucinous carcinoma, intraductal papilloma, mucocele, ductal carcinoma in situ | mucinous carcinoma, ductal carcinoma in situ, intraductal papilloma, columnar cell change without atypia, sclerosing adenosis |

**Table 5:** Error Analysis: Samples of erroneous predictions for disease severity and individual diagnoses.

To our knowledge, the proposed HCSBC model is one of first to classify 59 various pathology diagnosis in an end-to-end model NLP model and present the generalization on the two external datasets (MGH and Mayo Clinic). We also release a PathologyBERT - a pre-trained masked language model which was trained on 347,173 histopathology specimen reports and publicly released the language model in the Huggingface repository to trigger more model development in the breast cancer area. To support pathology informatics development and to enable the reproducibility of our experiments, we released the pre-training and fine-tuning code as well as our pre-trained representation models[29]. In addition, we developed a user-friendly graphical user interface (GUI)[1] for our system to upload data and obtain prediction results, and we publicly released in the Huggingface[2] repository[1]. We performed a through error analysis to highlight the consistent error pattern in our model model inference and showed that other than the 'Missed diagnosis', two other errors can be easily handled by pre-screening of the annotation and post-processing of the model output.

There are multiple potential applications of this model including populating tumor registries, identifying patient cohorts for research, conducting population health analyses, and deep learning dataset development, which was our primary use case. To develop deep learning models to identify breast cancer on mammography, a large number of images must first be labeled with their appropriate diagnoses. This would be an extremely laborious task if done manually for tens of thousands of patients. Instead, we used our model to automatically extract structured diagnoses labels from pathology reports and assign them to the appropriate mammograms for a patient. The results from this work have already been used to validate commercial and research breast cancer detection and risk prediction models[38].

While HCSBC shows significant promises in regards to pathology cancer diagnose extraction, it faces some limitations. Because we employ a hierarchical model, the accuracy of the diagnosis extraction depends upon on correct prediction of the disease severity category. Because the model is trained on pathology reports collected from a single institution from 2013 to 2020, it may underperform on older reports or at sites with different vocabulary and report structure. However, given the semi-structured nature of reporting, varying template structures may not pose a significant generalizability challenge and can also be addressed by pre-processing and section segmentation. The generalization shown by the model on two external dataset also proves the ability. Another limitation is the smaller input size of 64 tokens as compared to 512 for BERT and 128 for ClinicalBERT. This value works well with our current training corpus where we can easily extract a *Diagnosis* section with mean size of 42 tokens. However, this may pose problems when applying our model to a longer pathology reports from other institutions. In future, we intend to alleviate above-mentioned limitations by employing multi-institutional data for fine-tuning of our model.